# Using Connected Vehicle Trajectory Data to Evaluate the Effects of Speeding

Jorge Ugan, Mohamed Abdel-Aty, *Member, IEEE*, Zubayer Islam, *Member, IEEE*

*Abstract*—Speeding has been and continues to be a major contributing factor to traffic fatalities. Various transportation agencies have proposed speed management strategies to reduce the amount of speeding on arterials. While there have been various studies done on the analysis of speeding proportions above the speed limit, few studies have considered the effect on the individual's journey. Many studies utilized speed data from detectors, which is limited in that there is no information of the route that the driver took. This study aims to explore the effects of various roadway features an individual experiences for a given journey on speeding proportions. Connected vehicle trajectory data was utilized to identify the path that a driver took, along with the vehicle related variables. The level of speeding proportion is predicted using multiple learning models. The model with the best performance, Extreme Gradient Boosting, achieved an accuracy of 0.756. The proposed model can be used to understand how the environment and vehicle's path effects the drivers' speeding behavior, as well as predict the areas with high levels of speeding proportions. The results suggested that features related to an individual driver's trip, i.e., total travel time, has a significant contribution towards speeding. Features that are related to the environment of the individual driver's trip, i.e., proportion of residential area, also had a significant effect on reducing speeding proportions. It is expected that the findings could help inform transportation agencies more on the factors related to speeding for an individual driver's trip.

*Index Terms*—Probe vehicle data, connected vehicle data; speeding; machine learning

## I. INTRODUCTION

According to the National Highway Traffic Safety Administration (NHTSA), for more than two decades, speeding has been involved in approximately one-third of all motor vehicle fatalities. Speeding is also a major contributing factor for pedestrian and bicyclist fatalities on urban and suburban arterials [1, 2]. The ongoing speeding issue is an indication of the importance of speed management strategies in order to achieve vision zero. To be able to reduce the speeding related fatalities, it is important to understand the factors that are related to speeding.

Jorge Ugan is a PhD Candidate at the Department of Civil Environmental and Construction Engineering at University of Central Florida, FL 32826 (e-mail: jorgeugan@knights.ucf.edu).
Mohamed Abdel-Aty is a Professor at the Department of Civil Environmental and Construction Engineering at University of Central Florida, FL 32826 (e-mail: M.Aty@ucf.edu).
Zubayer Islam is a Postdoctoral Scholar at the Department of Civil Environmental and Construction Engineering at University of Central Florida, FL 32826 (e-mail: zubayer_islam@knights.ucf.edu).

### A. Probe Vehicle Data

The majority of speed related studies have utilized speed data from detectors. The data from these detectors can provide information at the specific location of the detector, but the data is limited for upstream and downstream of the detector location on a segment. To get a better understanding of the speed on a segment, studies have shown interest in probe data [3]. However, the validation studies of the probe data suggested that the speed data could be affected by factors such as signalized intersections and volume [3, 4]. To address this limitation, the speed of a vehicle's journey was adjusted by calculating the traversal speed of the individual's journey. Then, the vehicle's speed was compared against the speed limit, to determine the degree of speeding. In addition to the variables relating to speed, other variables where calculated related to the vehicle's journey, such as the time a vehicle spent stopped at an intersection, the number of intersections a vehicle passed through, the number of turn a vehicle took, etc. Two models were developed to analyze the factors relating to speeding, a model which predicts the degree of speeding at a specific point and a model which predicts the speeding of the overall vehicle journey.

### B. Roadway Characteristics

The Florida Department of Transportation (FDOT) developed a roadway classification system for planning, designing, and operating the state transportation network. The roadway classification system includes eight context classifications for all non-limited-access state roadways [5]. In urban and suburban areas, there are three major context classifications: (i) C3R which is the suburban residential road; (ii) C3C which is the suburban commercial road; and (iii) C4 which is the urban general road. The definition of the road context classification is based on the land use types and street patterns. In general, C3R is for mostly residential uses within large blocks and a disconnected or sparse roadway network, while C3C is for mostly non-residential uses within large building footprints and large parking lots within large blocks and a disconnected or sparse roadway network. C4 is a mix of uses set within small blocks with a well-connected roadway network and the roadway network usually connects to residential neighborhoods immediately along the corridor or behind the uses fronting the roadway. The specific context classification of the arterial which a vehicle was on was considered as a variable in the model. In addition, variables related to the roadway geometry was considered in the model, i.e., number of lanes, median width, etc.

In summary, factors relating to a vehicle's journey and the surrounding environment can influence speeding on arterials. However, the existing studies typically aggregate the vehicle probe data to model the proportion of speeding. These models do not consider the influence a vehicle's journey has on their speed choice, and the factors related. To fill the research gap, this paper aims to develop two speeding level models, (i) to determine the degree of speeding at a certain point and (ii) to determine the degree of speeding for a vehicle's journey. Speeding proportions were calculated based on the speed limit, and five speeding levels were calculated based on the speeding proportions. Other variables, such as hard braking, hard acceleration, time stopped at an intersection, number of turns, etc. from each vehicle journey are also used as input variables. Linear Discriminant Analysis (LDA), Support Vector Machine (SVM), Random Forest (RF), Gradient Boosting model (GBM), and Extreme Gradient Boosting (XGBT) are used. XGBT achieved the best experiment result.

## II. Literature Review

Various studies have explored the factors that are attributed to drivers' speed choice. Mahmud et al. analyzed the minimum, average, 75th percentile, and maximum speed for low-speed urban streets and identified the effects of road geometry, road environment, and traffic flow [6]. Bassani et al. analyzed the 85% percentile speed on urban arterials and identified the effects of road attributes such as the presence of shoulder, bus and taxi lane, and sidewalks on drivers' speed choice [7]. Bhowmik et al. and Cai et al. developed multilevel ordered probit fractional split models to analyze vehicle speed by different categories [8, 9]. The models analyzed the effects of roadway attributes, traffic data, land use, socio-demographic characteristics, and environmental factors on the speed proportions by different arterials. Eluru et al. analyzed the effects of various other roadway geometric factors on speed for arterials, including speed limit, number of lanes, lane width, and number of sidewalks [10]. Ghasemzadeh and Ahmed explored speeding behavior by using naturalistic driving data and developed a multilevel logistic model to capture the regional heterogeneity in speeding behavior [11]. Kong et al. and Yu et al. applied classification models to investigate the speeding behavior based on the naturalistic driving data [12, 13]. It was found that driving on roadways with low functional class, experience of congestion, and presence of median is associated with a higher speeding pattern. Another finding showed that when passing by houses or streets, drivers are more conservative with a least probability of speeding, and on road segments with a lower posted speed limit, drivers have a larger probability of speeding, especially when they suddenly enter an exceedingly speed-limited segment.

Speeding proportion could be a good measure to reflect the overall speeding level. Speed data with the speed limit on the roadway is needed to calculate the speeding proportions. Traditionally, transportation agencies and state departments of transportation install fixed sensors such as loop detectors and cameras for monitoring traffic. Although, the fixed sensors could provide relatively accurate traffic data, there is high cost associated with the deployment and maintenance of these sensors. These sensors are also limited to their geographical scalability as they need to be installed in a large number to determine the traffic condition in an area [14]. Mostly, they are installed on major freeways and critical arterials. In recent years, the probe vehicle technology has provided a cost-effective alternative for monitoring traffic and the data coverage has been growing significantly [3]. Several studies have been conducted to validate the accuracy and reliability of probe source data [3, 15-17]. It suggested that the data quality of probe vehicles improved significantly. However, the probe data still has limitations and the accuracy could be affected by factors such as traffic, speed limit, and signalized intersections. Arterials in urban areas with traffic signals experience a larger variability since vehicles stop during the red time [3]. Hence, these limitations must be considered when calculating the variables related to a vehicles journey.

In summary, this study aims to contribute to the literature of speeding analysis from the following three aspects: (1) explore the effects of features related to driving behavior, individual's journey, and roadway attributes to speeding; (2) explore various machine learning models to understand the effects of speeding and classify a vehicle's speeding level; (3) identify areas with high levels of speeding through the probe vehicle data. This paper is organized as follows: a brief discussion in earlier literature has been presented in this section. The data preparation section describes the data used for the analysis and proposes the method for adjusting the probe data. The methodology section presents the machine learning models for analyzing the speeding proportion. The modeling results of the speeding proportion levels are presented in results and discussions section. Finally, the last section concludes the findings of this paper.

## III. Data Preparation

The dataset used in this paper was provided by Wejo. It contains vehicle specific data from several manufacturers. It mainly has non-commercial fleet data which better represents the vehicles on the roadways. Instantaneous data is sent from the vehicle to the cloud in near real-time. The dataset consists of GPS location, heading, speed, postal code, journeyId and dataPointId. The sampling rate of the dataset was limited to 3 seconds. Two weeks of data was used in this study : 11th to 17th November 2019 and 9th to 16th November 2020. The entire dataset had a total of 273,010,898 GPS points.

The data preparation pipeline is shown in Fig. 1. It was filtered to limit it to Seminole County in Florida where there are over 350 signalized intersections along with a toll road SR 417 and part of the interstate I-4. The probe vehicle data was then augmented with several features such as speed limit, context classification, functional class, land use, roadway geometry and presence of intersections. Thus, the vehicle dynamic features along with geometric roadway attributes were present in the dataset. Finally, the dataset was aggregated per journey. A journey in the Wejo data is defined as the interval between an ignition start to an ignition stop. This resulted in over 900,000 journeys across the two weeks.

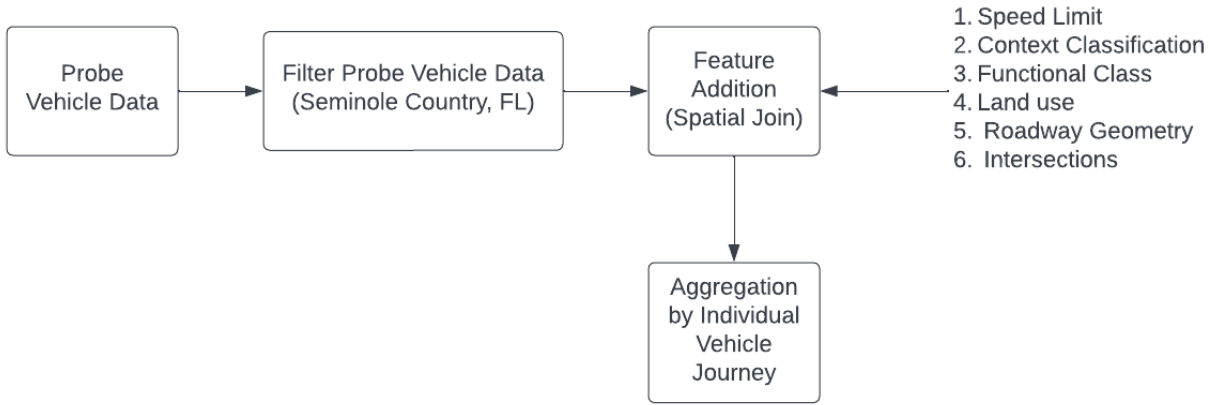

Fig. 1. Data Preparation Pipeline

The speeding proportion was calcucalted using the vehicle speed and the speed limit information as shown in (1).

$$y = \frac{speed - speed\ limit}{speed\ limit} \qquad (1)$$

The speeding proportions of the different roadway segments and a few hotspot locations are shown in Fig. 2. A segment is highligted in Fig. 2 if it has reported at least a thousand data points of any speeding porportion. It can give an understanding of the notable segments in Seminole County that experience speeding proportions between 20-40% above the speed limit. A few locations experience over 60% above the speed limit.

Table I shows the relationship between the speed proportion above the speed limit an individual travels and the speeding level. Any speed that was less than 5 % over the speed limit was considered to be non-speeding, and any speed greater than 80% was considered to be the highest speeding level (speeding level 5).

TABLE I
Relationship between the speeding proportion and speeding level

| Speeding Proportion | Speeding Level |
|---|---|
| (0.00, 0.05) | 0 |
| [0.05, 0.20) | 1 |
| [0.20, 0.40) | 2 |
| [0.40, 0.60) | 3 |
| [0.60, 0.80) | 4 |
| [0.80, 1.51) | 5 |

The descriptive statistics of the prepared feature data set and brief description is mentioned in Table II. The Pearson correlation coefficient of the features are shown in Fig. 3.

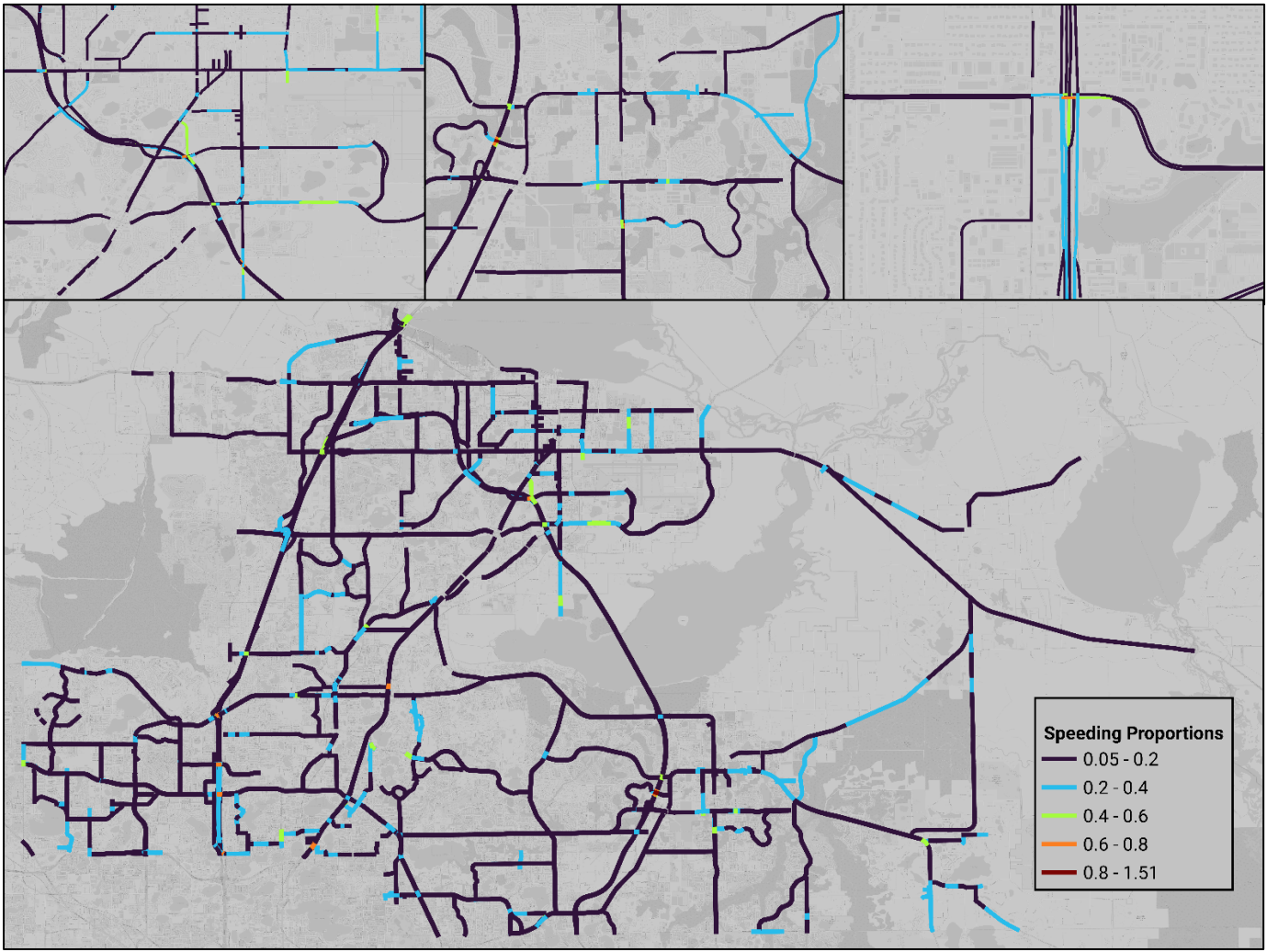

Fig. 2.  Speeding Proportions in Seminole County

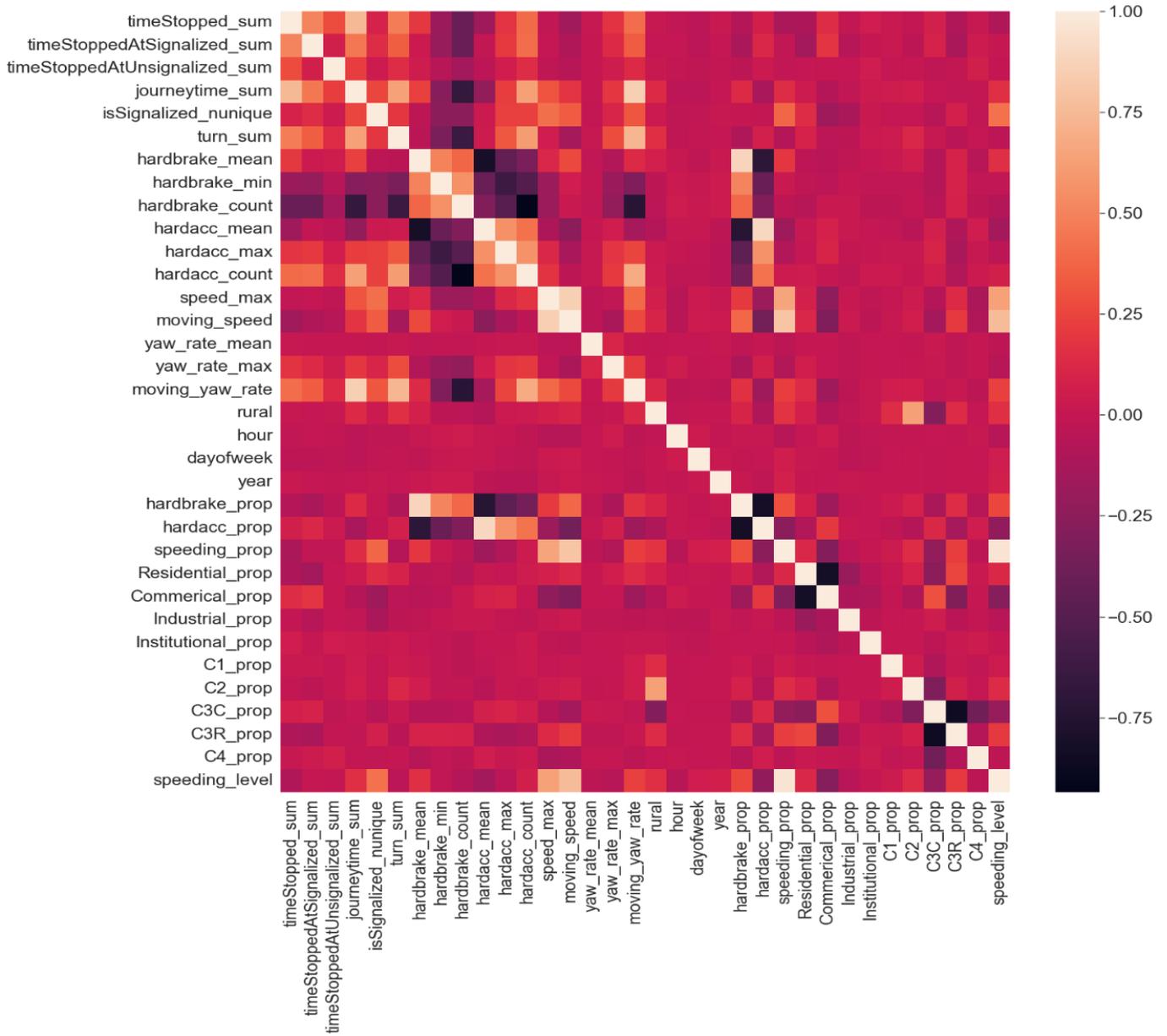

Fig. 3. Pearson correlation coefficient of the features

TABLE II
Descriptive Statistics of Aggregated Data

| Variable | Description | mean | std | min | max |
|---|---|---|---|---|---|
| timeStopped_sum | Time the vehicle spent stopped (secs) | 320.42 | 560.13 | 0 | 37744 |
| timeStoppedAt Signalized_sum | Time the vehicle spent stopped at Signalized Intersections (secs) | 102.99 | 179.94 | 0 | 16356 |
| timeStoppedAt Unsignalized_sum | Time the vehicle spent stopped at Unsignalized Intersections (secs) | 21.11 | 112.02 | 0 | 19969 |
| journeytime_sum | Total time of the vehicle's journey | 1038.63 | 866.27 | 1 | 39292 |
| isSignalized | Number of signalized intersections a vehicle passed through | 2.88 | 0.36 | 1 | 3 |
| turn_sum | Total number of turns in a vehicle's journey | 13.56 | 9.84 | 0 | 645 |
| hardbrake_mean | Average of hard brake in a vehicle's journey (m/s²) | -0.08 | 0.05 | -1.92 | 0 |
| hardbrake_min | Minimum hard brake in a vehicle's journey (m/s²) | -2.17 | 0.7 | -16.75 | 0 |
| hardbrake_count | Number of hard brakes in a vehicle's journey | -23.78 | 19.23 | -797.31 | 0 |
| hardacc_mean | Average of hard acceleration in a vehicle's journey (m/s²) | 0.07 | 0.04 | 0 | 0.91 |
| hardacc_max | Maximum hard acceleration in a vehicle's journey (m/s²) | 1.84 | 0.61 | 0 | 15.57 |
| hardacc_count | Number of hard accelerations in a vehicle's journey | 20.04 | 16.99 | 0 | 657.37 |
| moving_speed | Average speed, while vehicle in motion (mph) | 51.55 | 22.17 | 1.15 | 165.38 |
| yaw_rate_mean | Average angular velocity of the vehicle's journey (degrees per second) | 1.66 | 1.04 | 0 | 50.33 |
| yaw_rate_max | Maximum angular velocity of the vehicle's journey (degrees per second) | 23.7 | 7.77 | 0 | 174 |
| moving_yaw_rate | Average angular velocity of the vehicle's journey when the vehicle changes heading (degrees per second) | 154 | 101.25 | 0 | 3965 |
| hour | Hour the vehicle of traveling in the beginning of the journey | - | - | 0 | 23 |
| dayofweek | Day of the week the vehicle of traveling in the beginning of the journey | - | - | 0 | 6 |
| year | The year the vehicle was traveling in | - | - | 2019 | 2020 |
| hardbrake_prop | Proportion of hard braking in a vehicle's journey | 0.11 | 0.06 | 2.03 | 0 |
| hardacc_prop | Proportion of hard acceleration in a vehicle's journey | 0.09 | 0.06 | 0 | 1.28 |
| speeding_prop | Proportion of speeding in a vehicle's journey | 0.27 | 0.19 | 0 | 1.51 |
| Residential_prop | Proportion of residential area a vehicle traveled through | 0.34 | 0.26 | 0 | 1 |
| Commerical_prop | Proportion of commercial area a vehicle traveled through | 0.51 | 0.28 | 0 | 1 |
| Industrial_prop | Proportion of industrial area a vehicle traveled through | 0.03 | 0.08 | 0 | 1 |
| Institutional_prop | Proportion of institutional area a vehicle traveled through | 0.01 | 0.05 | 0 | 1 |
| C1_prop | Proportion of C1 roads a vehicle traveled through | 0 | 0.02 | 0 | 1 |
| C2_prop | Proportion of C2 roads a vehicle traveled through | 0.01 | 0.06 | 0 | 1 |
| C3C_prop | Proportion of C3C roads a vehicle traveled through | 0.83 | 0.28 | 0 | 1 |
| C3R_prop | Proportion of C3R roads a vehicle traveled through | 0.13 | 0.25 | 0 | 1 |
| C4_prop | Proportion of C4 roads a vehicle traveled through | 0.03 | 0.13 | 0 | 1 |
| speeding_level | Level of speeding for the vehicle's journey | 1.78 | 1.07 | 0 | 5 |

## IV. Methodology

To analyze levels of speeding proportions in Seminole County, Florida, we have gathered probe vehicle data for two weeks in 2019 and 2020. The data was aggregated by the journey of each individual, in total 490,456 journeys were obtained after cleaning. The dataset with the important variables were then split into training and testing datasets with a 70:30 ratio. Five different machine learning techniques (Support Vector Machine, Random Forest, GBM, LDA, XGBoost) were applied to the dataset, and the performances of the models were evaluated. The precision, recall, F1-score, and accuracy were used to assess the model performance. A tuning process was applied to determine the best set of parameters for each ensemble model, specifically the maximum depth of trees and number of trees to control the overfitting of the models. The testing results of the five models are summarized in Table III. It is clearly shown that XGBoost is able to provide significantly more accurate predictions. Hence, the trained XGBoost model will be used in the following analysis of feature effects.

### A. Random Forest (RF)

Random Forest (RF) is a tree-based classifier that employs two distinct machine learning techniques: random feature selection and bagging [18]. Random feature collection creates decision trees immediately, whereas bagging creates each tree separately. A decision tree model starts at the root node and split the data on the features that result in the largest information gain (IG). This partition process is repeated iteratively until the child node has values all belong to the same class [19]. Rather than employing all of the features in the decision trees, Random Forest selects the features of the subsets at random. For forecasting the output of a new dataset, Random Forest uses the mean value of the outputs from random independent bootstrap training data

### B. Support Vector Machine (SVM)

The Support Vector Machine (SVM) method is a supervised learning algorithm that is widely used. Given a data set D in the form of $\{x_i, y_i\}_{i=1}^{N}$, where $x_i \epsilon R_d$ are the samples, and the $y_i$ is the label, SVM maps the feature vector $x_i$ to a $N$-dimensional space, with $N$ as the number of features of the samples. For a multiclass classification problem, SVM breaks down the multiclassification problem into multiple binary classification problems finds the hyperplane (decision boundary) to distinctly separate the multiple classes of samples. The distance between the two classes, is regarded as margin distance. SVM uses a loss function to maximize the margin distance, which is to solve (2):

$$J = \min \frac{1}{2} w^T w + C \sum_{i=1}^{N} \varepsilon_i \qquad (2)$$

subject to:

$$y_i(w^T K_i(x_i) + b) \geq 1 - \varepsilon_i$$

where, $w$ is weight vector, $C$ is cost coefficient, and $\varepsilon_i$ is a slack variable for the non-separable data, and $K_i$ is the kernel function to transform data to the feature space. There are different kernel functions to use, such as linear functions, radial basis functions (rbf), etc.

### C. Linear Discriminant Analysis (LDA)

The objective of Linear discriminant analysis (LDA) is to find a linear combination of features that characterizes or separates two (or more) classes [20]. To do so, this algorithm maximizes the variability between the classes and reducing the variability within the classes.

### D. Gradient Boosting (GBM)

Gradient Boosting model (GBM) is a machine learning method that utilizes many decision trees (weak learners) to generate results. For GBM, at each iteration, a new tree is added. The subsequent trees will give extra weights to the samples that are incorrectly classified by the prior tree. Weighted voting is used to generate the final classification result based on all the trees [21].

### E. Extreme Gradient Boosting (XGBoost)

XGBoost (eXtreme Gradient Boosting) is an extension of the popular tree gradient boosting algorithms [22]. Boosting is the mechanism to add models recursively until optimal performance metrics are achieved. In gradient boosting, new models are added that predict the residuals of the prior which are added together for the final prediction. XGBoost has been proven to be an efficient and scalable version of gradient boosting trees that is also capable of utilizing maximum memory and hardware resources for data intensive models. It is also able to integrate sparsity aware data handling capabilities as well as a weighted quantile sketch for approximate learning as shown by Chen and Guestrin [23]. The authors were also able to generate a scalable package by gaining insights on cache access patterns, data compression and sharding that aided in training billions of samples in the meagerest of resources.

Mathematically XGBoost is summarized with (3), (4), and (5),

$$f = \omega_{q(x(t))} \qquad (3)$$

$$L^{(t)} = \sum_{i=1}^{n} l\left(y_i, \hat{y}_i^{t-1} + f_t(x_i)\right) + \Omega(f_t) \qquad (4)$$

$$\Omega(f_t) = \Upsilon T + 0.5 * \lambda \sum_{i=1}^{T} \omega_i^2 \qquad (5)$$

where, $x(t)$ is the input data, $y_i$ is the target variable while $\hat{y}_i$ is the predicted value for the i-th sample, $q$ is the function that maps a sample into a tree structure, $\omega$ is the leaf weight, $T$ is the number of leaf nodes, $L$ is the loss function, $\Upsilon$ and $\lambda$ are the regularization parameters.

### F. Model Performance Evaluation

A confusion matrix for multiclass classification is shown in Fig. 4. TN (true negative) corresponds to the number of actual negative samples that are correctly predicted. FP (false positive) corresponds to the number of actual negative samples that are wrongly predicted. FN (false negative) corresponds to the number of actual positive samples that are wrongly predicted. TP (true positive) corresponds to the number of samples in conflict class that are correctly predicted.

Fig. 4. A confusion matrix for multiclass classification

Using the above measures, the following metrics, precision, recall, F1-score, and the overall accuracy are calculated:

- Precision: the probability that 'A' speeding level in classification results are correctly classified.
- Recall: the probability that all 'A' speeding level in ground truth are classified as 'A'.
- Accuracy: the proportion of correctly classified samples among all the samples.
- F1-score: F1-score is estimated based on precision and recall, a high F1-score usually indicates the model's good overall performance.

$$Precision = \frac{TP}{TP + FP}$$

$$Recall = \frac{TP}{TP + FN}$$

$$F_1 = 2 \times \frac{(Precision \times Recall)}{(Precision + Recall)}$$

$$Accuracy = \frac{TP + TN}{TP + TN + FP + FN}$$

## V. Results and Discussion

Following feature selection, 32 variables were used as inputs to predict the speeding level. A training and testing ratio of 70:30 was maintained for the speeding level prediction. Table III shows the performances (Precision, Recall, F1-Score, and Accuracy) of all the machine learning models on the training dataset in predicting speeding level of an individual journey. As shown in Fig. 5, the overall accuracy of the training datasets varied from 60% to 76%, with XGBoost having the highest accuracy and SVM having the least. Except for the SVM model, all of the models demonstrated good precision in the prediction of the six speeding levels.

The recall performance of XGBoost was also the best for all speeding level prediction, except for two speeding level (speeding level 2 and 3). RF had a better recall for speeding level 2 and SVM had a better recall for speeding level 3. However, a detailed look into the confusion matrix showed that SVM did not accurately predict any journeys which had speeding level 2 and 4, other studies have also shown SVM not being able to accurately predict some class [24, 25]. It is clearly shown that the boosting methods could gain significantly better performance than the bagging method for estimating speeding levels. When comparing between two boosting methods, it could be found that XGBoost is able to provide slightly more accurate predictions.

These findings suggest that the model performance varies due to the variation of the dataset and the distribution of the speeding levels in the dataset.

The feature importance and the confusion matrix for the XGBoost model is shown in Fig. 5 and Fig. 6, respectively. For the feature importance, the top 20 most important features are displayed. Unsurprisingly, the average moving speed of the journey significantly prevails as the most important feature. The next most important features are related to the proportion of land use which the individual's journey took place. Features that are related to the driving behavior of the journey (i.e., yaw rate, hard brake, hard acceleration) also appeared to be very significant in the model.

The confusion matrix on the test data set is shown in Fig. 6. Around 76% of the journeys were correctly classified by their speeding level $\left(\frac{111,306}{147,137}\right)$ while 24% journeys are wrongly classified $\left(\frac{35,831}{147,137}\right)$. It is also important to note that out of the 24% of the speeding levels that were wrongly classified 91% $\left(\frac{32,589}{35,831}\right)$ of the journeys were classified into the neighboring speeding level (i.e., at most one speeding level above or below the true class). Furthermore, the XGBoost model is able to classify speeding level of the journey at 98% $\left(\frac{143,895}{147,137}\right)$ accuracy with one speeding level margin of error.

TABLE III
Modeling Results

| Model | Speeding Level | Metric | | | Accuracy | Support |
|---|---|---|---|---|---|---|
| | | Precision | Recall | F1-Score | | |
| SVM | 0 | **0.887** | 0.443 | 0.591 | 0.210 | 17099 |
| | 1 | 0.256 | 0.001 | 0.003 | | 40723 |
| | 2 | - | - | - | | 57203 |
| | 3 | 0.167 | **0.999** | 0.286 | | 23064 |
| | 4 | - | - | - | | 7413 |
| | 5 | 0.926 | 0.100 | 0.18 | | 1635 |
| RF | 0 | 0.866 | 0.712 | 0.782 | 0.739 | 17099 |
| | 1 | 0.704 | 0.690 | 0.697 | | 40723 |
| | 2 | 0.706 | **0.828** | 0.762 | | 57203 |
| | 3 | 0.762 | 0.623 | 0.685 | | 23064 |
| | 4 | **0.912** | 0.721 | 0.805 | | 7413 |
| | 5 | 0.968 | 0.839 | 0.899 | | 1635 |
| GBM | 0 | 0.849 | 0.744 | 0.793 | 0.755 | 17099 |
| | 1 | **0.719** | 0.707 | 0.713 | | 40723 |
| | 2 | 0.734 | 0.813 | 0.772 | | 57203 |
| | 3 | 0.768 | 0.685 | 0.724 | | 23064 |
| | 4 | 0.884 | 0.791 | 0.835 | | 7413 |
| | 5 | 0.930 | 0.874 | 0.901 | | 1635 |
| LDA | 0 | 0.709 | 0.601 | 0.651 | 0.601 | 17099 |
| | 1 | 0.588 | 0.539 | 0.562 | | 40723 |
| | 2 | 0.611 | 0.740 | 0.670 | | 57203 |
| | 3 | 0.572 | 0.431 | 0.491 | | 23064 |
| | 4 | 0.530 | 0.364 | 0.431 | | 7413 |
| | 5 | 0.358 | 0.778 | 0.491 | | 1635 |
| XGBT | 0 | 0.851 | **0.745** | **0.795** | **0.756** | 17099 |
| | 1 | 0.718 | **0.708** | **0.713** | | 40723 |
| | 2 | **0.735** | 0.813 | 0.772 | | 57203 |
| | 3 | **0.768** | 0.689 | **0.726** | | 23064 |
| | 4 | 0.903 | **0.795** | **0.846** | | 7413 |
| | 5 | 0.961 | **0.881** | **0.920** | | 1635 |

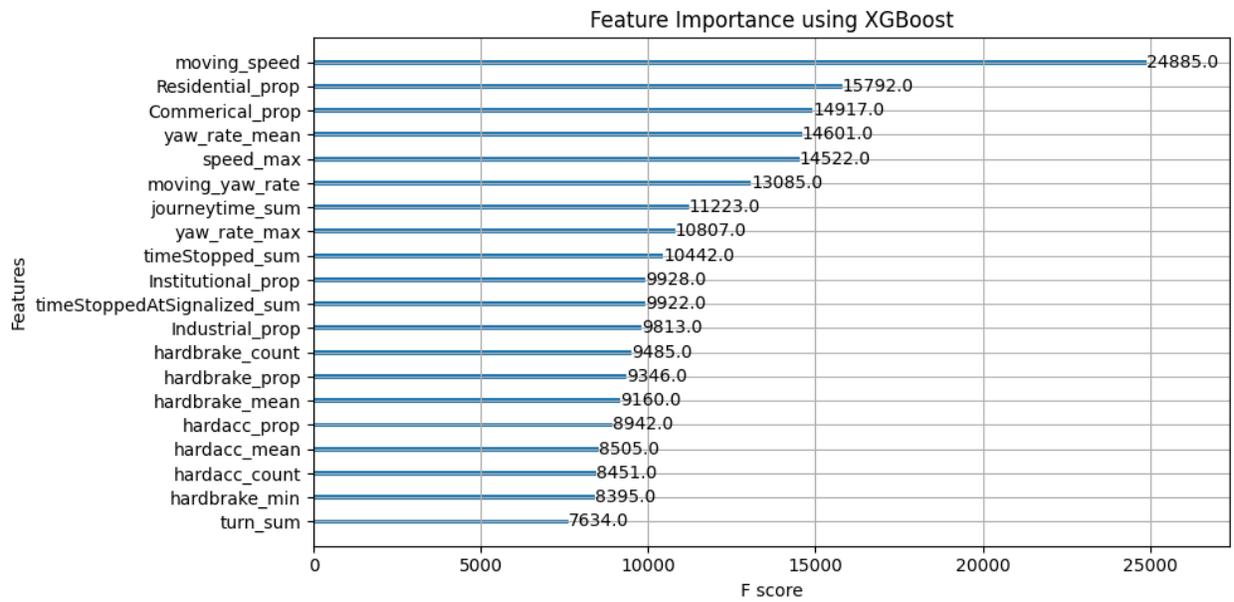

Fig. 5. Feature importance from XGBoost model

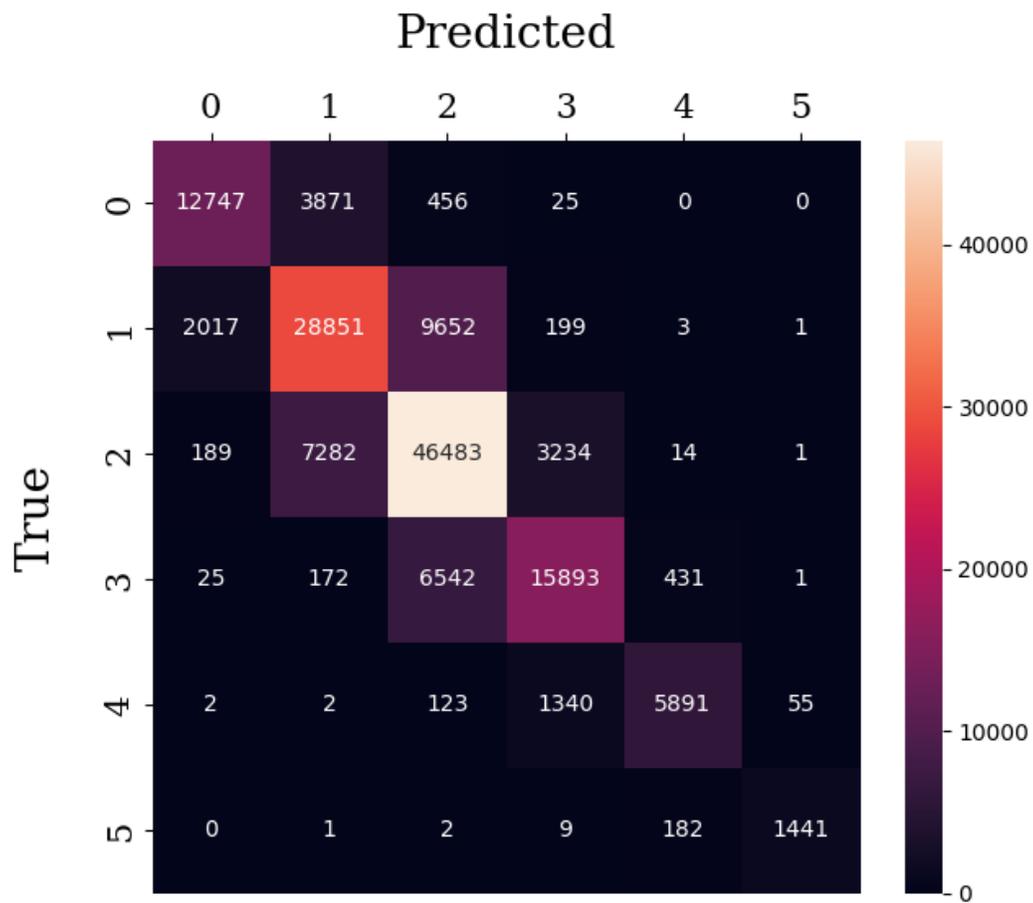

Fig. 6. Confusion matrix for the speeding levels using XGBoost

The SHapley Additive exPlanations (SHAP) values is a method based on cooperative game theory and used to increase transparency and interpretability of machine learning models [26]. SHAP values were calculated for the features in every observation in the test dataset, and their values will disclose the individual contribution of each speeding level. The positive SHAP value means positive impact on prediction for the speeding level, leading the model to predict towards that specific speeding level. Negative SHAP value means negative impact, leading the model to predict towards another speeding level.

Furthermore, to illustrate the model's capacity of distinguishing between the speeding levels, t-Distributed Stochastic Neighbor Embedding (t-SNE) is introduced to visualize features for each speeding level. t-SNE is a powerful tool for high-dimensional data visualization [27, 28]. It maps multi-dimensional data to two or more dimensions suitable for human observation. Specifically, t-SNE of the extracted features from the SHAP values is shown in Fig. 7. It can be shown that the model can distinguish the level of speeding accurately. In each of the six graphs of Fig. 7, there is a clear separation of among the various speeding levels. Fig. 7(a) shows the likelihood of a journey belonging to speeding level 0. For this journeys that are most likely non-speeding are grouped together, and the journeys which are speeding are also group together in different sections of the graph. The distinction of speeding levels 1, 2, and 3 is not as clear as speeding level 0 in Fig. 7 (b), (c), and (d), but the likelihoods of being in the speeding level or not are still distinguish fairly well. The distinction of speed level 4 and 5 are clear in Fig. 7 (e) and (f), that the higher likelihood of a journey being part of those speeding level are grouped together. As expected, when comparing Fig. 7(a) and Fig. 7(f), the likelihoods for each journey are almost opposite, since this is comparing non-speeding with the highest speeding level (speeding level 5).

Four dependence plots were done on features related to an individual's journey were, all the features were among the top 20 most important features. The best fitted polynomial functions from degrees from 1 to 5 were applied to each speeding class in the data and the model with the best fit plotted in the Fig.s. As shown in Fig. 8(a), the average speed of the journey is an important feature for determining the likelihood of a speeding level. The four interior speeding levels (speeding level 1, 2, 3, 4) show the maximum of the curves increasing in as the average speed increases. This goes on to say that as the average increases the more likely the journey is to be in a higher speeding level. Speeding level 0 shows the opposite, as expected, as the average speed of a journey increases the less likelihood of non-speeding. Fig. 8(b) shows the effect of amount of time stopped at a signalized intersection on speeding level. The four curves to the top portion of the curve are all high speeding levels (speeding level 2, 3, 4, and 5) and the bottom curves are the non-speeding level and lower speeding level (speeding level 0 and 1). This goes on to say that as an individual's stops more at intersections, their likelihood of speeding also increases. Many studies have also similar findings, where individuals speed more to make up for lost time. Ellison and Greaves utilized naturalistic driving data to test time savings from speeding, and also found that perceived time savings is often a motivation for speeding [29]. Fig. 8(c) shows effect of the proportion of hard braking for an individual's journey. Focusing on the 0.20 to 0.50 proportion of hard brake in an individual's journey, it is shown that the two highest speeding levels (speeding level 4 and 5) increase at a high rate. This can be interpreted as an individual applies more hard brake, the more likely they are speeding at a high level. Hu et al. also analyzed the effect of hard brakes on speeding, the study also concluded that higher hard brakes were shown in vehicles with higher average speeds [30]. Fig. 8(d) shows a similar interpretation to Fig. 8(c), focusing on the 0.40 to 0.50 proportion of hard acceleration, it can be seen that the higher speeding levels are higher and the lower speeding levels curves.

The two dependence plots that were done on features related to the land use of the journey, were also among the top 20 most important features. Fig. 9(a) shows the effect of the proportion of residential area of an individual's journey. Focusing on the higher proportion, 0.8 to 1, of residential area in an individual's journey, it is shown that lower speeding levels (non-speeding level and speeding level 1) are higher than the high speeding levels (speeding levels 2, 3, 4, and 5). With the increase of an individual driving in a residential area the more likely they are to be at a low speeding level or non-speeding. This can be attributed to many factors related to residential areas (i.e., lower speed limits, higher number of pedestrians, etc.). Fig. 9(b) as a similar interpretation to Fig. 9(a), where the non-speeding and lower speeding level curves are higher than the high speeding levels.

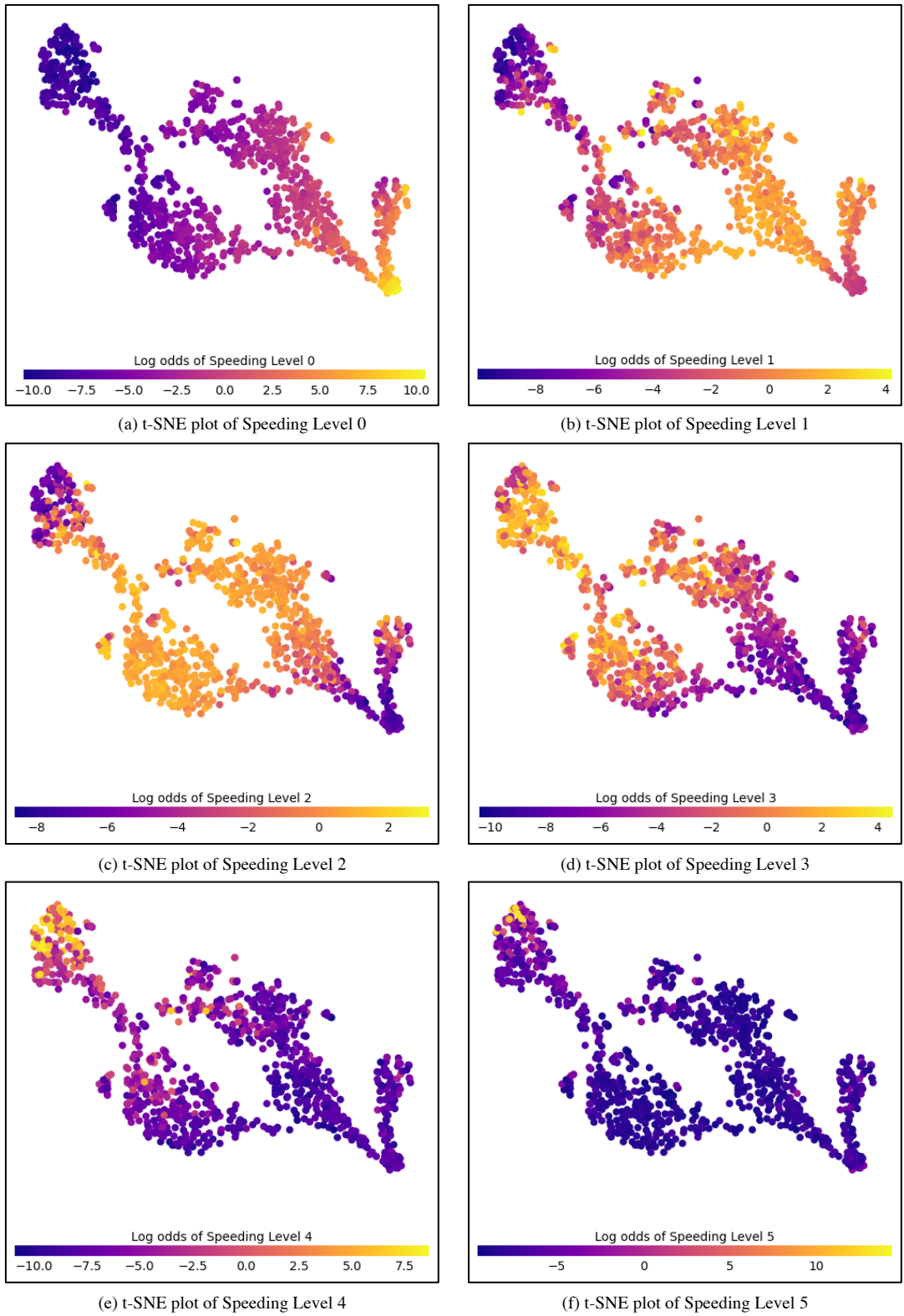

(a) t-SNE plot of Speeding Level 0

(b) t-SNE plot of Speeding Level 1

(c) t-SNE plot of Speeding Level 2

(d) t-SNE plot of Speeding Level 3

(e) t-SNE plot of Speeding Level 4

(f) t-SNE plot of Speeding Level 5

Fig. 7. t-SNEs of extracted features using SHAP values

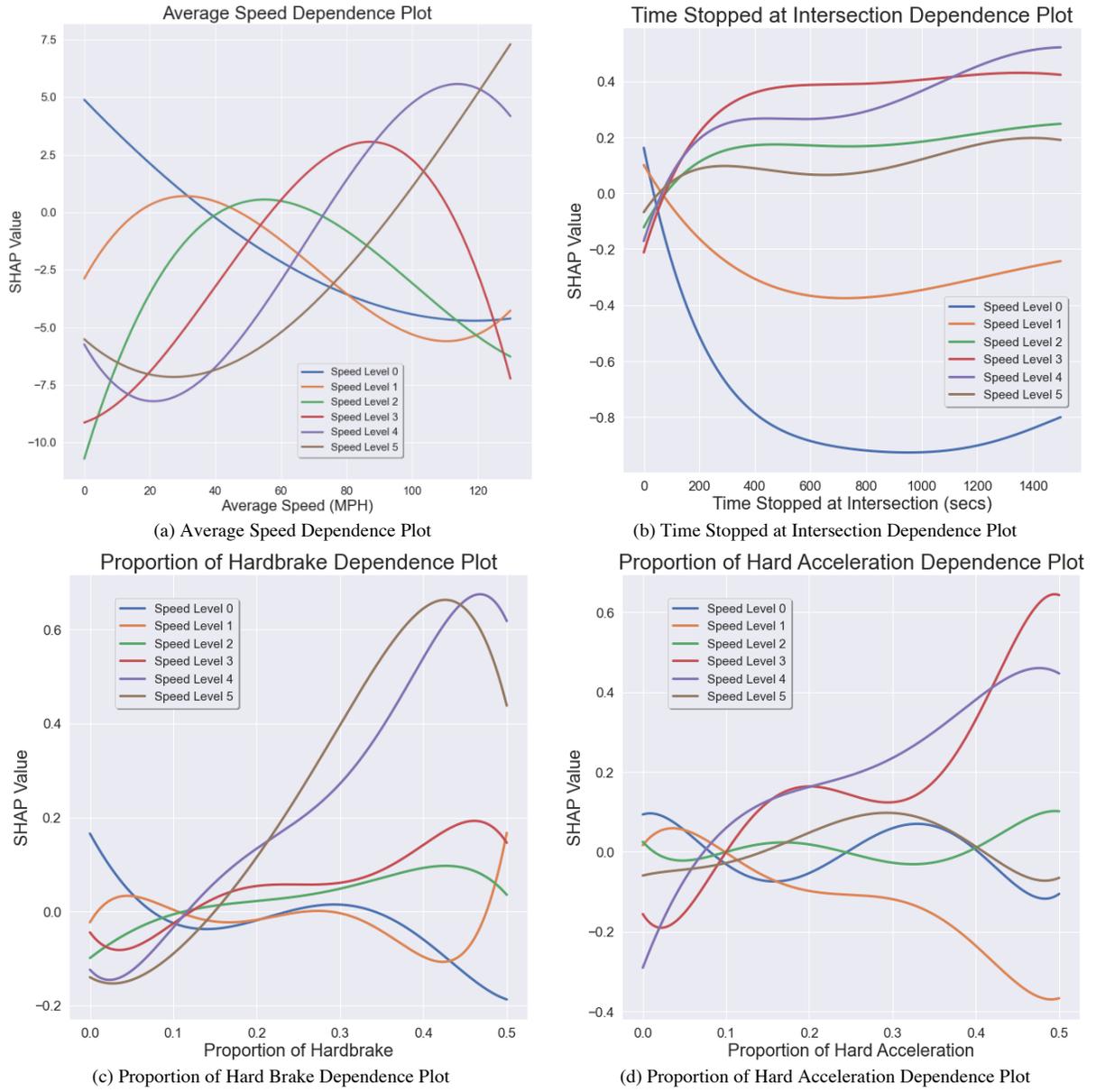

(a) Average Speed Dependence Plot

(b) Time Stopped at Intersection Dependence Plot

(c) Proportion of Hard Brake Dependence Plot

(d) Proportion of Hard Acceleration Dependence Plot

Fig. 8. Dependence plots for features related to an individual's journey

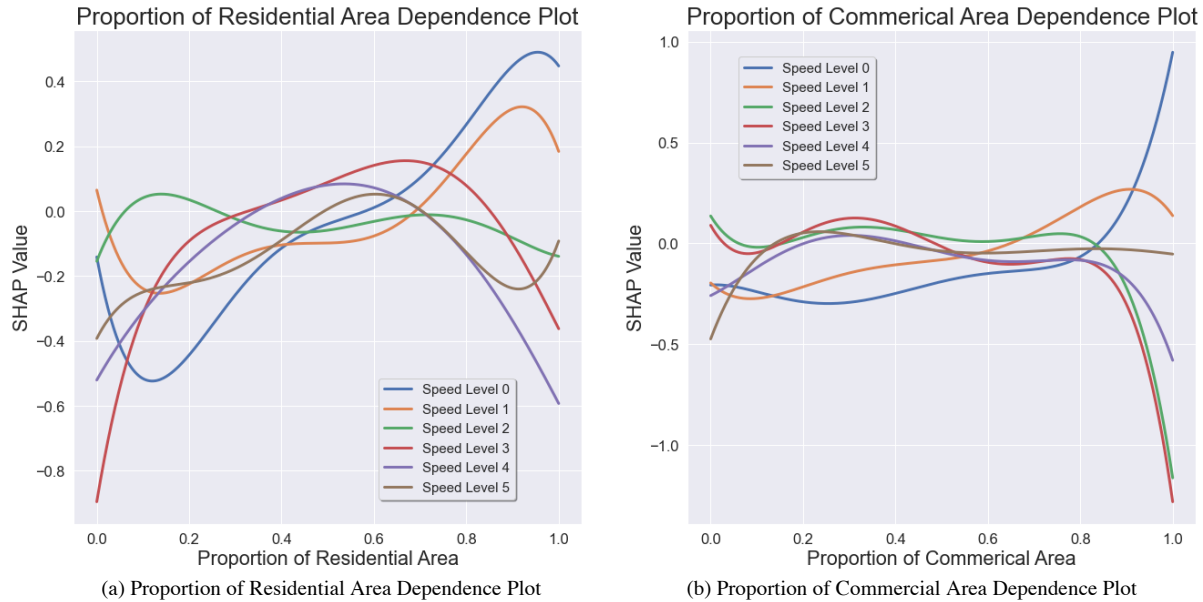

(a) Proportion of Residential Area Dependence Plot

(b) Proportion of Commercial Area Dependence Plot

Fig. 9. Dependence plots for features related to land use of the journey

## VI. CONCLUSIONS

This study applied different machine learning methods to explore the effects of an individual's journey on speeding. Around 273,000,000 GPS points were obtained from Seminole County, Florida. These data points were fused with data from the surrounding environment (roadway features, land use, etc.). The data was then aggregated to each individual's journey, which resulted in around 490,000 individual journeys. The aggregated data set included information based on the route the individual took in addition to the probe vehicle data, such as the time spent at intersections, number of intersections passed, number of turns, proportion of residential area, proportion of commercial area, etc. The proportion of various features were calculated such as the proportion of hard brake, proportion of acceleration, etc. to get an understanding of how these proportion effect an individual in speeding.

Five machine learning models (i.e., Support Vector Machine, Random Forest, GBM, LDA, XGBoost) were applied to estimate the speeding level of an individual's journey. The comparison results suggested that the XGBoost could provide the best fit. The explainable machine learning method was used to explore the effects of the features related to an individual's journey on the speeding. Factors including average speed, hard brake proportions, hard acceleration proportions, road attributes, land use characteristics were examined. The result revealed that features related to an individual journey (i.e., the specific route the individual took) are very important contributing factors for determining the level of speeding. It was suggested that the more time spent stopped at signalized intersections the more likely the individual would speed a high level throughout the journey. The reason behind this is due to the individual's belief in saving time by speeding. The features related to acceleration and deceleration were also significant to determine an individual's speeding level. As the number of hard brake and hard acceleration increased the more likely the individual would speed throughout the journey. It is important to note that the hard brake proportions had a clearer distinction between the lower speeding levels and higher speeding levels than the hard acceleration proportions. In addition, the land use proportion which the journey passed through also proved to be important in determining the speeding level. At higher proportions of residential and commercial areas, there was less speeding. Overall, this paper succeeds in verifying the possibilities to distinguish between various levels of speeding for an individual's journey. The results of this paper can be used for the implementation of an advanced traffic management system, which has the potential to reduce speeding. Transportation agencies will be able to know the features related to the population of the high-speed drivers and where they are originating from, which will allow for safer transportation planning.

## VII. ACKNOWLEDGMENT

The authors also acknowledge Wejo for providing the connected vehicle data. This paper and its contents, including conclusions and results, are solely those of the authors; they do not represent opinions or policies of Wejo.

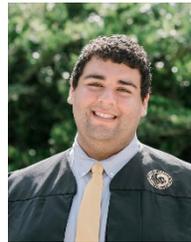

**Jorge Ugan** received the B.Sc. and M.Sc. degrees in transportation engineering from the University of Central Florida, Orlando, Florida, in 2020 and 2021, respectively. He is currently pursuing the Ph.D. degree with the Civil Engineering Department, University of Central Florida. His research interests include traffic safety analysis, pedestrian and bicycle safety, and intelligent transportation systems.

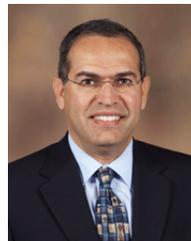

**Mohamed Abdel-Aty** (Member, IEEE) is currently a Pegasus Professor and the Chair of the Civil, Environmental, and Construction Engineering Department, University of Central Florida (UCF), Orlando, FL, USA. He has published more than 620 articles (325 in journals). He received nine best paper awards from ASCE, TRB, and WCTR. His main expertise and interests are in the areas of ITS, traffic safety, simulation, CAV, and active traffic management.

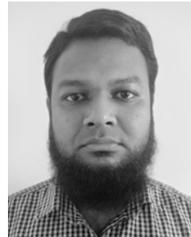

**Zubayer Islam** received the B.Sc. in electrical and electronics engineering from Bangladesh University of Engineering and Technology, Dhaka, Bangladesh, in 2017, and the Ph.D. degree in transportation engineering from the University of Central Florida, Orlando, FL, USA, in 2021. He is currently a Postdoctoral Scholar of transportation engineering with the University of Central Florida.